\documentclass[11pt]{article}
\usepackage{eacl2017}
\usepackage{times}
\usepackage{url}
\usepackage{latexsym}
\usepackage{times}
\usepackage{helvet}
\usepackage{courier}
\usepackage{multirow}
\usepackage{changepage}
\usepackage{color}
\usepackage{caption}
\usepackage{subcaption}
\usepackage{graphicx}
\usepackage{footnote}
\usepackage{balance}
\usepackage{arydshln}

\eaclfinalcopy 
\def\eaclpaperid{78} 

\title{Detecting (Un)Important Content\\ for Single-Document News Summarization \thanks{*Accepted By EACL 2017}}

\author{Yinfei Yang \\
Redfin Inc. \\
Seattle, WA 98101 \\
\normalsize{\tt yangyin7@gmail.com} \\\And
Forrest Sheng Bao \\
University of Akron\\ Akron, OH 44325 \\
\normalsize{\tt forrest.bao@gmail.com} \\\And
Ani Nenkova \\
University of Pennsylvania \\
Philadelphia, PA 19104 \\
\normalsize{\tt nenkova@seas.upenn.edu} \\}

\date{}

\begin{document}
\maketitle
\begin{abstract}
We present a robust approach for detecting intrinsic sentence importance in news, by training on two corpora of document-summary pairs. When used for single-document summarization, our approach, combined with the ``beginning of document'' heuristic, outperforms a state-of-the-art summarizer and the beginning-of-article baseline in both automatic and manual evaluations. These results represent an important advance because in the absence of cross-document repetition, single document summarizers for news have not been able to consistently outperform the strong beginning-of-article baseline.
 
\end{abstract}

\section{Introduction}

To summarize a text, one has to decide what content is important and what can be omitted. 
With a handful of exceptions \cite{DBLP:conf/emnlp/SvoreVB07,Berg-Kirkpatrick:2011:JLE:2002472.2002534,DBLP:conf/uai/KuleszaT11,cao2015ranking,DBLP:conf/acl/0001L16}, modern summarization methods are unsupervised, relying on on-the-fly analysis of the input text to generate the summary, without using indicators of intrinsic importance learned from previously seen document-summary pairs. 
This state of the art is highly unintuitive, as it stands to reason that some aspects of importance are learnable. Recent work has demonstrated that indeed supervised systems can perform well without sophisticated features when sufficient training data is available \cite{DBLP:conf/acl/0001L16}. 

In this paper we demonstrate that in the context of news it is possible to learn an accurate predictor to decide if a sentence contains content that is \emph{summary-worthy}.
We show  that the predictors built in our approach are remarkably consistent, providing almost identical predictions on a held out test set, regardless of the source of training data. 
Finally we demonstrate that in single-document summarization task our predictor, combined with preference for content that appears at the beginning of the news article, results in a summarizer significantly better than a state-of-the-art global optimization summarizer. The results hold for both manual and automatic evaluations. 

In applications, the detector of unimportance that we have developed can potentially improve snippet generation for news stories, detecting if the sentences at the beginning of the article are likely to form a good summary or not. This line of investigation was motivated by our previous work showing that in many news sub-domains the beginning of the article is often an uninformative teaser which is not suitable as an indicative summary of the article \cite{Yang2014}.  



\section{Corpora}
\label{sec:dataset}

One of the most cited difficulties in using supervised methods for summarization has been the lack of suitable corpora of document-summary pairs
where each sentence is clearly labeled as either important or not~\cite{zhou2003web}. 
We take advantage of two currently available resources: archival data from the Document Understanding Conferences (DUC) \cite{Over2007} and the New York Times (NYT) corpus (\url{https://catalog.ldc.upenn.edu/LDC2008T19}). The DUC data contains document-summary pairs in which the summaries were produced for research purposes during the preparation of a shared task for summarization. The NYT dataset contains thousands such pairs and the summaries were written by information scientists working for the newspaper.

DUC2002 is the latest dataset from the DUC series in which annotators produced extractive summaries, consisting of sentences taken directly from the input. 
DUC2002 contains 64 document sets.
The annotators created two extractive summaries for two summary lengths (200 and 400 words), for a total of four extracts per document set.
In this work, a sentence from the original article that appears in at least one of the human extracts is labeled as \emph{important} (summary-worthy).
All other sentences in the document are treated as \textit{unlabeled}. Unlabeled sentences could be truly not summary-worthy but also may be included into a summary by a different annotator \cite{Nenkova:2007:PMI:1233912.1233913}. We address this possibility in Section~\ref{sec:method}, treating the data as partially labeled. 




For the NYT corpus, we work with 19,086 document-summary pairs published between 1987 and 2006 from the Business section.

Table \ref{tab:example} in Section 5 shows a summary from the NYT corpus. These are abstractive, containing a mix of informative sentences from the original article along with abstractive re-telling of the main points of the article, as well as some meta-information such as the type of article and a list of the photos accompanying the article. It also shows the example of lead (opening) paragraph along with the summary created by the system we propose, InfoFilter, with the unimportant sentence removed.

In order to label sentences in the input, we employee Jacana ~\cite{jacana} for word alignment in mono-lingual setting for all pairs of article-summary sentences.
A sentence from the input is labeled as \textit{important} (summary-worthy) if the alignment score between the sentence and a summary sentence is above a threshold, which we empirically set as 14 based on preliminary experiments.
All other sentences in the input are treated as \emph{unlabeled}. 
Again, an unlabeled sentence could be positive or negative.


\section{Method}
\label{sec:method}
As mentioned earlier, existing datasets contain clear labels only for positive sentences. 
Due to the variability of human choices in composing a summary, unlabeled sentences cannot be simply treated as negative. 
For our supervised approach to sentence importance detection, a semi-supervised approach is first employed to establish labels.


\subsection{Learning from Positive and Unlabeled Samples}
Learning from 
positive (e.g., \textit{important} in this paper) and unlabeled samples 
can be achieved by the methods proposed in~\cite{Lee03,Elkan2008}.
Following~\cite{Elkan2008}, 
we use a two-stage approach to train a detector of sentence importance from positive and unlabeled examples.

Let $y$ be the importance prediction for a sample,
where $y=1$ is expected for any positive sample and $y=0$ for any negative sample.
Let $o$ be the ground-truth labels obtained by the method described in Section~\ref{sec:dataset}, 
where $o=1$ means that the sentence is labeled as positive (important) and $o=0$ means unlabeled.

In the first stage, we build an estimator $e$, equal to the probability that a sample is predicted as positive given that it is indeed positive,  $p(o=1|y=1)$. 
We first train a logistic regression (LR) classier with positive and unlabeled samples, treating the unlabeled samples as negative.
Then $e$ can be estimated as $\Sigma_{x\in P}(LR(x)/|P|)$, where $P$ is the set of all labeled positive samples,
and $LR(x)$ is the probability of a sample $x$ being positive, as predicted by the LR classifier.
We then calculate $p(y=1|o=0)$ using the estimator $e$, the probability for an unlabeled sample to be positive as:
$w=\frac{LR(x)}{e} / \frac{1-LR(x)}{1-e}$. 
A large $w$ means an unlabeled sample is likely to be positive, whereas a small $w$ means the sample is likely to be negative. 

In the second stage, a new dataset is constructed from the original dataset.
We first make two copies of every unlabeled sample,
assigning the label $1$ with weight $w$ to one copy and the label $0$ with weight $1-w$ to the other.
Positive samples remain the same and the weight for each positive sample is $1$.
We call this dataset the \textit{relabeled data}. 

We train a SVM classifier with linear kernel on the relabeled data. This is our final detector of important/unimportant sentences.

\subsection{Features}
The classifiers for both stages use dictionary-derived features which indicate the types / properties of a word, along with several general features.

\paragraph{MRC}
The MRC Psycholinguistic Database \cite{MRC88} is 
a collection of word lists with associated word attributes according to judgements by multiple people.
The degree to which a word is associated with an attribute is given as a score within a range.
We divide the score range into 230 intervals. The number of intervals was decided empirically on a small development set and was inspired by prior work of feature engineering for real valued scores \cite{klebanov2013using}.
Each interval corresponds to a feature; the value of the feature is the fraction of words in a sentence whose score belongs to this interval.  
Six attributes are selected: imagery, 
concreteness, 
familiarity, 
age-of-acquisition, 
and two meaningfulness attributes. 
In total, there are 1,380 MRC features. 


\paragraph{LIWC}
LIWC is a dictionary  that groups words in different categories, such as positive or negative emotions, self-reference etc.
and other language dimensions relevant in the analysis of psychological states.
Sentences are represented by a histogram of categories, indicating the percentage of words in the sentence associated with each category.
We employ LIWC2007 English dictionary which contains 4,553 words with 64 categories.

\paragraph{INQUIRER}
The General Inquirer \cite{INQUIRER} is another dictionary of 7,444 words, grouped in 182 general semantic categories. 
For instance, the word \textit{absurd} is mapped to tags NEG and VICE.  
Again, a sentence is represented with the histogram of categories occurring in the sentence.

\paragraph{General}
We also include features that capture general attributes of sentences including:
\textit{
total number of tokens, 
number of punctuation marks, 
if it contains exclamation marks,
if it contains question marks, 
if it contains colons,
if it contains double quotations.  
}

\section{Experiments on Importance Detection}

We train a classifier separately for the DUC2002 and the NYT 1986-2006 corpora. 
The DUC model is trained using the articles and summaries from DUC2002 dataset, where 1,833 sentences in total appear in the summaries. 
We also randomly sample 2,200 non-summary sentences as unlabeled samples to balance the training set. 
According to the criteria described in NYT corpus section, there are 22,459 (14.1\%) positive sentences selected from total of 158,892 sentences.
Sentences with Jacana alignment scores less than or equal to 10 form the unlabeled set, including 20,653 (12.9\%) unlabeled sentences in total. 
Liblinear~\cite{liblinear} is used for training the two-stage classifiers.

\subsection{Test Set}

The test set consists of 1,000 sentences randomly selected from NYT dataset for the year 2007.
Half of the sentences are from the Business section, where the training data was drawn.
The rest are from the U.S. International Relations section (\textit{Politics} for short), to test the stability of prediction across topic domains.
Three students from the University of Akron
annotated if the test sentences contain important summary-worthy information.

For each test (source) sentence from the original article, 
we first apply Jacana to align it with every sentence in the corresponding summary. 
The summary sentence with the highest matching score is picked as the target sentence for the source sentence.
Each pair of source and target sentences is presented to students and they are asked to mark if the sentences share information.
Sentences from the original article that contribute content to the most similar summary sentence are marked as positive;
those that do not are marked as negative.
The pairwise annotator agreements are all above $80\%$ and the pairwise Kappa ranges from $0.73$ to $0.79$.

The majority vote becomes the label of the source (article) sentence.
Table~\ref{tab:nyt_details} presents the distribution of final labels. The classes are almost balanced, with slightly more negative pairs overall.


\begin{table}[!htbp]
\caption{The distribution of the annotated labels}
\centering
\small
\begin{tabular}{c | c | c }
   Section  & Positive & Negative  \\
\hline 
Business  & 232 (46.4\%) & 268 (53.6\%) \\
Politics  & 219 (43.8\%) & 281 (56.2\%) \\
Total     & 451 (45.1\%) & 549 (54.9\%) \\  \hline
\end{tabular}
\label{tab:nyt_details}
\end{table}

\subsection{Evaluation Results}

In the process above, we have obtained a set of article sentences that contribute to the summary (positive class) or not (negative class)\footnote{We assume that an article sentence not contributing to the summary does not contribute any content to the summary sentence that is closest to the article sentence.}.

Table~\ref{tab:nyt_res} shows the evaluation results on the human-annotated test set.
The baseline is assuming that all sentences are summary-worthy.
Although the unimportant class is the majority (see Table~\ref{tab:nyt_details}), predicting all test samples as not summary-worthy is less useful in real applications because we cannot output an empty text as a summary.

Each row in Table~\ref{tab:nyt_res} corresponds to a model trained with one training set.
We use dictionary features to build the models, i.e., NYT Model and DUC Model.
We also evaluate the effectiveness of the general features by excluding it from the dictionary features, 
i.e. NYT w/o general and DUC w/o general.
Precision, recall and F-1 score are presented for all models.
Models trained on the NYT corpus and DUC corpus are both significantly better than the baseline, with $p<0.0001$ for McNemara's test.
The NYT model is better than DUC model  overall according to F-1.
The results also show a noticeable performance drop when general features are removed.

We also trained classifiers with bag of words (BOW) features for NYT and DUC respectively,
i.e. BOW-NYT and BOW-DUC.
The classifiers trained on BOW features still outperform the baseline but are not as good as the dictionary and general sentence properties models.

\setlength\dashlinedash{0.2pt}
\setlength\dashlinegap{1.5pt}
\setlength\arrayrulewidth{0.3pt}

\begin{table}[!htbp]
\caption{Evaluation results on human annotations}
\centering
\small
\begin{tabular}{c | c | c | c}
           & Precision & Recall & F-1 \\
\hline 
NYT Model  & 0.582 & 0.846 & 0.689 \\
DUC Model  & 0.541 & 0.903 & 0.676 \\
\hdashline
NYT w/o General  & 0.547 & 0.847 & 0.664 \\
DUC w/o General  & 0.508 & 0.906 & 0.651 \\
\hdashline
BOW-NYT    & 0.520 & 0.852 & 0.645 \\ 
BOW-DUC    & 0.501 & 0.828 & 0.623 \\ 
\hdashline
Baseline   & 0.464 & 1.000 & 0.621 \\  \hline
\end{tabular}
\label{tab:nyt_res}
\end{table}

\subsection{NYT Model vs. DUC Model}
Further, we study the agreement between the two models in terms of prediction outcome. 
First, we compare the prediction outcome from the two models using NYT2007 test set. The Spearman's correlation coefficients between the outputs from the two models is around $0.90$, showing that our model is very robust and independent of the training set. 

Then we repeat the study on a much larger dataset, using
articles from the DUC 2004 multi-document summarization task. There are no single document summaries in that year but this is not a problem, because we use the data simply to study the agreement between the two models, i.e., whether they predict the same summary-worthy status for sentences, not to measure the accuracy of prediction.
There are 12,444 sentences in this dataset.
The agreement between the two models is very high ($87\%$) for both test sets.
Consistent with the observation above, the DUC model is predicting intrinsic importance more aggressively.
Only for a handful of sentences the NYT model predicts positive  (important) while the DUC model predicts negative (not important). 

We compute Spearman's correlation coefficients between the posterior probability for sentences from the two models.
The correlation is around $0.90$, indicating a great similarity in the predictions of the two models.

\section{Summarization}
\label{sec:summ}

\begin{table*}[!htbp]
\begin{adjustwidth}{-0.4cm}{}
\small
\caption{Example of unimportant content in the opening paragraph of an article. The detected unimportant sentences are italicized. The third panel shows a new summary, with unimportant content skipped.}
\centering
\begin{tabular}{|p{16.5cm}|}
\hline
\textbf{Human Summary: }
Pres Bush and his aides insist United States is committed to diplomatic path in efforts to stop Iran's suspected nuclear weapons program and support for terrorism, but effort is haunted by similar charges made against Iraq four years ago.  Democrats see seizure of Iranians in Iraq and attempts to starve Iran of money to revitalize its oil industry as hallmarks of administration spoiling for fight. some analysts see attempt to divert attention from troubles in Iraq .  administration insiders fear Bush's credibility has been deeply damaged.  Bush's advisors debate how forcefully to push confrontation with Iran.
\\
\hline
\textbf{Lead paragraph: }
\textit{This time, they insist, it is different.}
As President Bush and his aides calibrate how directly to confront Iran, they are discovering that both their words and their strategy are haunted by the echoes of four years ago, when their warnings of terrorist activity and nuclear ambitions were clearly a prelude to war.
\textit{''We're not looking for a fight with Iran,'' R. Nicholas Burns, the under secretary of state for policy and the chief negotiator on Iranian issues, said in an interview, just a few hours after Mr. Bush had repeated his warnings to Iran to halt ''killing our soldiers''} ...
\\
\hline
\textbf{New summary; unimportant sentences removed: }
As President Bush and his aides calibrate how directly to confront Iran, they are discovering that both their words and their strategy are haunted by the echoes of four years ago, when their warnings of terrorist activity and nuclear ambitions were clearly a prelude to war. 
Mr. Burns, citing the president's words, insisted that Washington was committed to ''a diplomatic path'', even as it executed a far more aggressive strategy, seizing Iranians in Iraq and attempting to starve Iran of the money it needs to revitalize a precious asset, its oil industry. 
Mr. Burns argues that those are defensive steps ...
\\
\hline
\end{tabular}
\label{tab:example}
\end{adjustwidth}
\end{table*}

We propose two importance-based approaches to improving single-document summarization.

In the first approach, \textbf{InfoRank}, the summary is constructed solely from the predictions of the sentence importance classifier. 
Given a document,
we first apply the sentence importance detector on each sentence to get the probability of this sentence being intrinsically important.
Then we rank the sentences by the probability score to form a summary within the required length.

The second approach, \textbf{InfoFilter}, uses the sentence importance detector as a pre-processing step. 
We first apply the sentence importance detector on each sentence, in the order they appear in the article. We keep only sentences predicted to be summary-worthy as the summary till the length restriction. This combines the preference for sentences that appear at the beginning of the article but filters out sentences that appear early but are not informative.

\subsection{Results on Automatic Evaluation}

The model trained on the NYT corpus is used in the experiments here. 
Business and politics articles (100 each) with human-generated summaries from NYT2007 are used for evaluation. 
Summaries generated by summarizers are restricted to $100$ words.
Summarizer performance is measured by ROUGE-1 (R-1) and ROUGE-2 (R-2) scores~\cite{rouge}. 

Several summarization systems are used for comparison here, including
LeadWords, which picks the first 100 words as the summary;
RandomRank, which ranks the sentences randomly and then picks the most highly ranked sentences to form a 100-word summary;
and Icsisumm~\cite{Icsisumm}, a state-of-the-art multi-document summarizer~\cite{Kai2014}.

Table~\ref{tab:nyt_summ} shows the ROUGE scores for all summarizers.
InfoRank significantly outperforms Icsisumm on R-1 score and is on par with it on R-2 score.
Both InfoRank and Icsisumm outperform  RandomRank by a large margin.
These results show that the sentence importance detector is capable of identifying the summary-worthy sentences.

LeadWords is still a very strong baseline single-document summarizer.
InfoFilter achieves the best result and greatly outperforms the LeadWords in both  R-1 and R-2 scores.
The $p$ value of Wilcoxon signed-rank test is less than $0.001$, indicating that the improvement is significant.
Table \ref{tab:example} shows the example of lead paragraph along with the InfoFilter summary with the unimportant sentence removed.

\begin{table}[!htbp]
\begin{adjustwidth}{-0cm}{}
\caption{Performance comparison on single-document summarization ($\%$)}
\centering
\small
\begin{tabular}{c |c | c || c |c |c }
System  & R-1 & R-2 & System & R-1 & R-2 \\
\hline
\textbf{InfoRank}   & \textbf{37.6} & 15.9 & \textbf{InfoFilter} & \textbf{50.7} & \textbf{30.2} \\ 
Icsisumm          & 33.3 & \textbf{16.0} & LeadWords         & 48.0 & 27.5 \\ 
RandomRank        & 31.9 & 8.7  \\ 
\end{tabular}
\label{tab:nyt_summ}
\end{adjustwidth}
\end{table}

The InfoFilter summarizer is similar to the LeadWords summarizer, but it removes any sentence predicted to be unimportant and replaces it with the next sentence in the original article that is predicted to be summary-worthy.
Among the 200 articles, 116 have at least one uninformative sentence removed.
The most frequent number is two removed sentences. There are 17 articles for which more than three sentences are removed. 

\subsection{Results on Human Evaluation}


We also carry out human evaluation, to better compare the relative performance of the  LeadWords and InfoFilter summarizers.
Judgements are made for each of the 116 articles in which at least one sentence had been filtered out by InfoFilter. 
For each article, we first let annotators read the summary from the NYT2007 dataset and then the two summaries generated by LeadWords and InfoFilter respectively. 
Then we ask annotators if one of the summary covers more of the information presented in the NYT2007 summary. The annotators are given the option to indicate that the two summaries are equally informative with respect to the content of the NYT summary.
We randomize the order of sentences in both LeadWords and InfoFilter summaries when presenting to annotators.

The tasks are published on Amazon Mechanical Turk (AMT) and each summary pair is assigned to 8 annotators. 
The majority vote is used as the final label.
According to human judgement, InfoFilter generates better summaries for 55 of the 116 inputs; for 39 inputs, the LeadWords summary is judged better. 
The result is consistent with the ROUGE scores, showing  that InfoFilter is the better summarizer.


\section{Conclusion}
In this paper, we presented a detector for sentence importance and demonstrated that it is robust regardless of the training data. The importance detector greatly outperforms the baseline.  Moreover, we tested the predictors on several datasets for summarization. In single-document summarization, the ability to identify unimportant content allows us to significantly outperform the strong lead baseline. 

\balance
\bibliography{eacl2017}
\bibliographystyle{eacl2017}

\end{document}